\PassOptionsToPackage{table}{xcolor}
\documentclass[10pt,twocolumn,letterpaper]{article}

\usepackage{cvpr}              
\usepackage{graphicx}
\usepackage{amsmath}
\usepackage{amssymb}
\usepackage{booktabs}
\usepackage{makecell}
\usepackage{times}
\usepackage{epsfig}
\usepackage{wrapfig}    
\usepackage{enumitem}
\usepackage{multirow}
\usepackage{algorithmic}
\usepackage{algorithm}
\usepackage{diagbox}    
\usepackage{booktabs}
\usepackage{caption}    
\usepackage{subcaption} 
\usepackage{bm}

\definecolor{mygray}{gray}{0.9}
\definecolor{comment}{RGB}{0, 102, 0}
\definecolor{myred}{named}{black}

\definecolor{cvprblue}{rgb}{0.21,0.49,0.74}
\usepackage[pagebackref,breaklinks,colorlinks,allcolors=cvprblue]{hyperref}

\def\paperID{4} 
\def\confName{CVPR}
\def\confYear{2025}

\title{Aligned Contrastive Loss for Long-Tailed Recognition}

\author{%
 \textbf{Jiali Ma}\textsuperscript{1} \quad 
 \textbf{Jiequan Cui}\textsuperscript{2} \quad 
 \textbf{Maeno Kazuki}\textsuperscript{3} \quad 
 \textbf{Lakshmi Subramanian}\textsuperscript{1}\\ 
 \textbf{Karlekar Jayashree}\textsuperscript{1} \quad 
\textbf{Sugiri Pranata}\textsuperscript{1} \quad 
 \textbf{Hanwang Zhang}\textsuperscript{2}\\
\small \textsuperscript{1}Panasonic R\&D Center Singapore\quad \textsuperscript{2}Nanyang Technological University\quad 
\textsuperscript{3}Panasonic Connect Co., Ltd. R\&D Division\\
\tt\small jiali.ma@sg.panasonic.com \quad jiequancui@gmail.com \quad maeno.kazuki@jp.panasonic.com\\
\tt\small lakshmi.subramanian@sg.panasonic.com \quad karlekar.jayashree@sg.panasonic.com \\
\tt\small sugiri.pranata@sg.panasonic.com \quad hanwangzhang@ntu.edu.sg\\}

\begin{document}
\maketitle
\begin{abstract}
In this paper, we propose an Aligned Contrastive Learning (ACL) algorithm to address the long-tailed recognition problem. Our findings indicate that while multi-view training boosts the performance, contrastive learning does not consistently enhance model generalization as the number of views increases. Through theoretical gradient analysis of supervised contrastive learning (SCL), we identify gradient conflicts, and imbalanced attraction and repulsion gradients between positive and negative pairs as the underlying issues. Our ACL algorithm is designed to eliminate these problems and demonstrates strong performance across multiple benchmarks. We validate the effectiveness of ACL through experiments on long-tailed CIFAR, ImageNet, Places, and iNaturalist datasets. Results show that ACL achieves new state-of-the-art performance.
\end{abstract}    
\section{Introduction}
\label{sec:intro}

Long-tailed recognition presents a critical challenge in the realm of computer vision due to the severely imbalanced distribution of different classes. 
With traditional classification methods, models trained on long-tailed data exhibit extremely imbalanced performance. In particular, they underperform in the underrepresented tail classes.
The resulting bias significantly impacts the fairness and efficacy of deep learning models in real-world applications, such as autonomous driving, face recognition on minority groups, and medical diagnosis of rare conditions. 

In recent years, contrastive learning has emerged as a promising paradigm for learning good representations in a self-supervised manner. Supervised contrastive learning (SCL)~\cite{khosla2020supervised} further extends self-supervised InfoNCE loss~\cite{oord2018representation} by incorporating label information. To address long-tailed recognition, PaCo~\cite{cui2021parametric}, GPaCo~\cite{cui2023generalized}, BCL~\cite{zhu2022balanced} and ProCo~\cite{du2024probabilistic} integrate SCL with logit compensation loss~\cite{menon2020long}, enabling both representation learning and rebalancing in a unified framework.
The effectiveness of SCL hinges critically on both the quality and quantity of positive pairs. Recent studies~\cite{cui2023generalized, zhu2022balanced, suh2023long, du2024probabilistic, li2022targeted, wang2021contrastive} have explored various strategies for defining positive pairs, including augmented views, same-class samples, and class-specific weights in the classifier head. These methods typically require large batch sizes or momentum queues to ensure sufficient positive and negative pairs. However, large batch sizes demands substantial GPU memory while offering only a marginal increase in positive pairs, and outdated features in momentum queues may introduce fluctuations and lead to inconsistencies in the learning process.

\begin{figure}[t!]
    \centering 
    \includegraphics[width=0.75\linewidth]{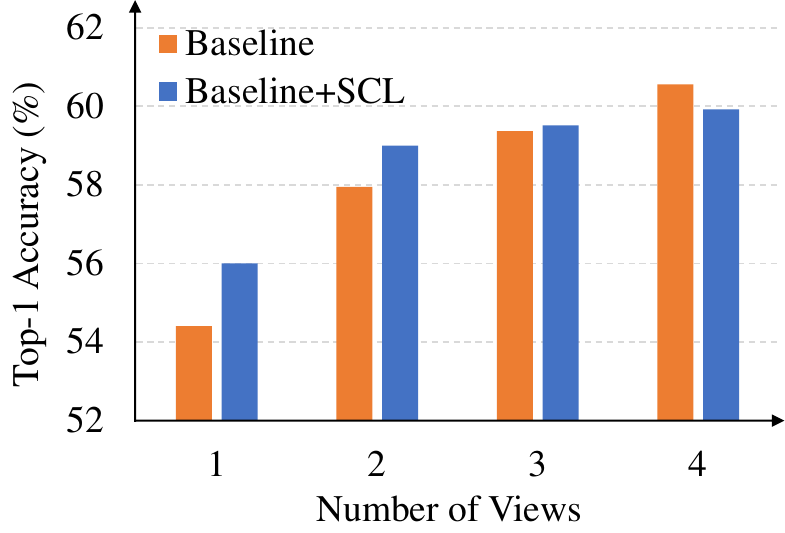} 
    \vspace{-2mm}
    \caption{Top-1 accuracy (\%) of Balanced Softmax~\cite{ren2020balanced} baseline (ResNeXt-50 backbone) with various number of views on ImageNet-LT dataset. We observe that multi-view training boosts the performance of long-tailed recognition while contrastive learning fails to continuously enhance performance due to gradients conflict and imbalanced attraction and repulsion gradients issues.} 
    \label{fig:view}
    \vspace{-5mm}
\end{figure}

To populate contrastive pairs, it’s intuitive to include multiple augmented views of the same instance as positives, as shown in~\cite{chen2020big,chen2021empirical,caron2020unsupervised}. 
Increasing the number of views leads to a quadratic growth in positive pairs, enhancing intra-class compactness and improving model performance. Furthermore,~\cite{fort2021drawing} shows that higher augmentation multiplicity also boosts accuracy in conventional classification losses.

\begin{figure}[t!]
    \centering 
    \includegraphics[width=0.8\linewidth]{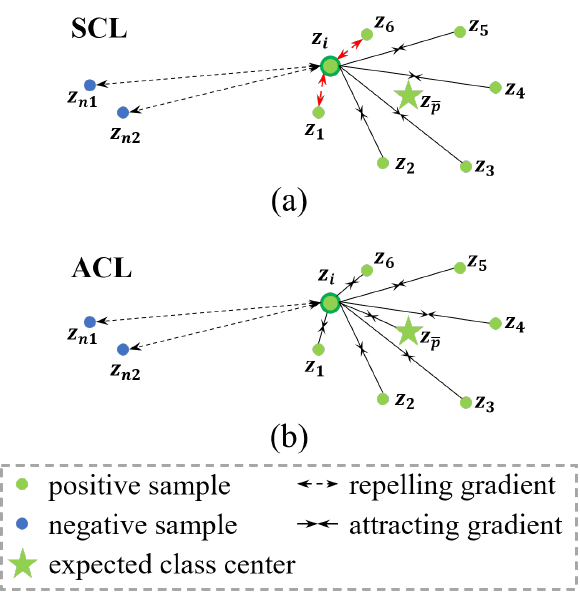} 
    \vspace{-3mm}
    \caption{\textcolor{myred}{Comparison between SCL and ACL. (a) In SCL, the training sample $z_i$ exerts repulsive forces on easy positive samples $z_1$ and $z_6$ due to their closer proximity compared to the averaged class center $z_{\bar{p}}$, \eg $distance(z_i, z_{1,6}) < distance(z_i, z_{\bar{p}})$.
    This repulsion (highlighted in red) introduces conflicting gradients and degrades model performance (see Section~\ref{sec:3.3} for detailed analysis). (b) Our proposed ACL mitigates these conflicting gradients by ensuring consistent attraction among all positive samples and the class center, promoting a more compact representation space.}}
    \label{fig:scl_acl}
    \vspace{-4mm}
\end{figure}
In this paper, we investigate the application of SCL and multi-view training in long-tailed scenarios, aiming to fully leverage their potential. 
As depicted in Fig.~\ref{fig:view}, training with an increasing number of augmented views on ImageNet-LT consistently improves the top-1 accuracy of the baseline model using Balanced Softmax loss, demonstrating the effectiveness of multi-view training. However, when SCL is incorporated, additional views do not always yield performance gains and may even cause degradation. This observation motivates a deeper exploration of the interplay between SCL and multi-view training.

Through theoretical analysis of pairwise gradient components in SCL, we identify an inherent conflict among gradients from different positive pairs. 
This conflict arises because SCL, similar to Softmax classification loss, encourages the alignment of a given positive pair while simultaneously pushing away all other pairs, including other potential positive samples from the same class.
From an instance-level perspective, the aggregated gradients from all positive pairs exert a repulsive force against easy positives that are closer to the current training sample than the expected class center. As shown in Fig.~\ref{fig:scl_acl} (a), the positive samples $z_1$ and $z_6$ experience repulsion from the training sample $z_i$ due to their proximity relative to the averaged class center $z_{\bar{p}}$, \eg $distance(z_i, z_{1,6}) < distance(z_i, z_{\bar{p}})$.
This effect can significantly impede representation learning, and we posit that the conflicting gradient intensifies as the number of positives increases, such as with additional augmented views.


In this work, we address the gradient conflict in SCL under multi-view training setting for long-tailed recognition.
We propose the aligned contrastive loss (ACL), which eliminates the conflicting positive terms and ensures consistent attraction among all positive pairs as shown in Fig.~\ref{fig:scl_acl} (b). 
ACL re-balances the gradient distribution between attraction and repulsion by re-weighting the negatives based on inverse class frequency. 
We validate the effectiveness of ACL on popular long-tailed benchmarks including CIFAR-LT, ImageNet-LT, iNaturalist 2018, and Places-LT, achieving state-of-the-art performance.

Our main contributions are summarized as follows.
\begin{itemize}
    \item Through theoretical pairwise gradient analysis of SCL, we identify an inherent gradient conflict between different positive pairs. The conflict intensifies as the number of positives increases in multi-view training (Section~\ref{sec:3.3}). 
    \item We propose ACL to alleviate the gradient conflict and unify consistent attraction among all positive pairs. ACL re-balances the attraction and repulsion gradients by re-weighting negative pairs, thus fully leveraging the benefits of multi-view training (Section~\ref{sec:5}).
    \item Extensive experiments on long-tailed recognition benchmarks demonstrate the superiority of our method across various tasks. We achieve new state-of-the-art (SOTA) results, \ie, 61.1\% on ImageNet-LT, 75.6\% on iNaturalist 2018 and \textcolor{myred}{42.4\% on Places-LT dataset} (Section~\ref{sec:6.3}).
\end{itemize}

\section{Related work}
\label{sec:related}

\subsection{Long-tailed recognition}
In long-tailed recognition, the class imbalance is traditionally addressed through re-balancing techniques. These include re-sampling, which over-samples minority classes or under-samples majority classes~\cite{buda2018systematic,byrd2019effect,drummond2003c4,pouyanfar2018dynamic,shen2016relay}, and re-weighting, which assigns inverse-frequency weights to classes in loss computation~\cite{cui2019class,huang2016learning,wang2017learning,khan2017cost, shu2019meta}. Both methods promote balanced classifier learning yet unexpectedly damage the representative ability of the deep features~\cite{kang2019decoupling, zhou2020bbn,zhang2023deep,nam2023decoupled}. Therefore, re-balancing strategies are usually used together with a 2-stage training paradigm.

Kang \etal~\cite{kang2019decoupling} proposed a decoupled training strategy for long-tailed recognition, where representation learning and classification are trained separately. 
BBN~\cite{zhou2020bbn} utilizes a bilateral branch network to dynamically balance features from instance-balanced and reversed sampling branches.

An alternative approach to long-tailed recognition involves adjusting logit values based on logarithmic label frequencies. Balanced Softmax~\cite{ren2020balanced} is introduced to address bias in Softmax loss. 
Menon \etal~\cite{menon2020long} further introduces post-hoc logit adjustment and a logit-adjusted classification loss, shifting from empirical risk minimization to balanced error minimization. 
This technique has been extensively adopted as a complementary enhancement in various long-tailed algorithms ~\cite{cui2021parametric,cui2023generalized,zhu2022balanced,suh2023long,zhu2024generalized}.

\subsection{Contrastive learning}
Contrastive learning has gained widespread adoption in self-supervised learning to enhance representation robustness by contrasting positive and negative pairs with augmented views~\cite{he2020momentum, chen2020simple,grill2020bootstrap,chen2021exploring}. SCL~\cite{khosla2020supervised} extends it to the supervised setting by encouraging distinctions between samples at the class level.
Recently, contrastive learning has become prevalent in long-tailed recognition~\cite{cui2023generalized, zhu2022balanced,suh2023long,du2024probabilistic,zhang2023deep, zhang2022fairness,hou2023subclass}.
KCL~\cite{kang2020exploring} integrates balanced feature space and cross-entropy classification discriminability using K positives.
TSC~\cite{li2022targeted} further aligns class features closer to target features on regular simplex vertices. 

Several works combine the merits of contrastive learning with logit adjustment techniques. 
PaCo~\cite{cui2021parametric} and GPaCo~\cite{cui2023generalized} seamlessly integrate these methods into a single loss, introducing parametric learnable class centers to expand contrastive pairs. 
BCL~\cite{zhu2022balanced} introduces class weight embeddings for comparison and adopts class averaging to balance the positive and negative pairs.
GML~\cite{suh2023long} creates class-wise queues for contrast samples and conducts knowledge distillation based on the features of a pre-trained teacher model.

In this work, we theoretically analyze the pairwise gradient of SCL and find that gradient conflict in positive pairs hinders effectiveness of the learning process in multi-view setting. Our proposed ACL eliminates the conflicting items to promote consistent attraction for all the positives.

\subsection{Augmentation multiplicity}
Multiple augmented views have emerged as a significant enhancement in contrastive learning frameworks to learn more robust and invariant representations.
The usage of multiple positive pairs is explored in works~\cite{chen2020big,chen2021empirical} and is proved to promote model performance with additional augmented views.
SwAV~\cite{caron2020unsupervised} introduces a multi-crop strategy, utilizing both global and local crops to enforce consistency across different image views. 
Additionally, the work~\cite{fort2021drawing} shows that increasing the multiplicity of augmentations improves accuracy in conventional classification losses.
Our work extends multi-view training to long-tailed recognition, proposing ACL to fully leverage the benefits of multiple views.
\section{Gradient conflict in SCL}
\label{sec:prelim}

\subsection{Preliminaries}
\label{sec:3.1}
Given an imbalanced training dataset $\mathcal{D}={\left\{ x_{i}, y_{i}\right\}}^{n}_{i=1}$, let $N_{j}$ denote the number of samples in the $j$-th class, $j\in\left\{1,2,...,C \right\}$. The distribution of $N_{j}$ follows a long-tailed pattern, \ie, $N_1 \geqslant  N_2 \geqslant ... \geqslant N_C$.
The task of long-tailed recognition is to learn a function mapping $\varphi$ from the image space $X$ to the target space $Y$. 
Specifically, $\varphi$ can be divided into a feature extractor $f: X \to Z \in R^{h}$ and a classifier $W: Z \to Y$, where $h$ is the feature dimension. 
Prior works focus on learning a balanced classifier or enhancing representation learning with data augmentations and contrastive learning. In this work, we further refine the representation learning by aligning and re-balancing contrastive learning across different pairs.

\subsection{SCL}
\label{sec:3.2}
SCL extends contrastive loss to supervised learning by contrasting positive and negative pairs, where positive pairs belong to the same class and negative pairs come from different classes~\cite{khosla2020supervised}.
 For sample $x_i$ with representation $z_i$, SCL is defined as
\begin{equation}
    \begin{aligned}
        \mathcal{L}_{i} = &\frac{1}{|P(i)|}\sum_{p \in P(i)}\mathcal{L}_{(i,p)} \\
        =&\frac{1}{|P(i)|}\sum_{p \in P(i)}-\mathrm{log}\frac{e^{z_i\cdot z_p/\tau}}{\sum_{a \in A(i)}^{}e^{ z_i \cdot z_a/\tau }},
    \end{aligned}
    \label{eq:scl}
\end{equation}
where $z_p$ represents the features of a same-class positive pair, $A(i)$ and $P(i)$ represent the sets of indices for all remaining samples and positive samples excluding instance $i$, and $\tau$ is the temperature parameter. 
As shown in Eq.~\eqref{eq:scl}, different positives occupy distinct positions.
We designate the positive $z_p$ in the numerator as the \textbf{\textit{effective positive}}, and refer to the remaining positives $z_j$ in the denominator ($j \in P(i)$ and $j \neq p$) as \textbf{\textit{non-effective positives}}.

Furthermore, Eq.~\eqref{eq:scl} highlights the synergy between SCL and Softmax classification loss, as both simultaneously minimize intra-class distances while maximizing inter-class separations in the feature space. We interpret SCL as the average Softmax loss over all positive samples, where representation $z_a$ serves as the weight vector in the $|A(i)|$-way classification layer. The pairwise loss can be formulated as
\begin{equation}
    \begin{aligned}
        \mathcal{L}_{(i,p)}=-\mathrm{log}\frac{e^{f_p}}{\sum_{a \in A(i)}^{}e^{f_a}}.
    \end{aligned}
    \label{eq:softmax}
\end{equation}
Here $f_a=z_i\cdot z_a/\tau$ is the logit of the $a$-th class. Eq.~\eqref{eq:softmax} implies that each pairwise contrastive loss works as a single-label classification with a unique ground-truth class, effectively classifying $z_{i}$ into the same class as $z_p$ among $|A(i)|$ alternatives. 
While optimizing $\mathcal{L}_{(i,p)}$ maximizes the ground-truth logit $f_p$, it simultaneously minimizes the logits of all other classes $a\in A(i), a \neq p$, including those of non-effective positives. This process generates a repulsive force against these non-effective positives, resulting in conflicting gradients.

\subsection{Pairwise gradient analysis}
\label{sec:3.3}
To fully investigate the conflicting gradients, we analyze the gradient of pairwise loss $\mathcal{L}_{(i,p)}$.
\begin{equation}
    \begin{aligned}
        \left. \frac{\partial \mathcal{L}_{(i,p)}}{\partial z_{k}} \right|_{k \in A(i)}
        =\frac{1}{\tau} \times 
        \begin{cases} 
        -(1-q_{(i, p)})z_i, & \text{if } k = p; \\ 
        q_{(i,k)}{z_i}, & \text{otherwise}.
        \end{cases}
    \end{aligned}
    \label{eq:grad_ind}
\end{equation}
where $q_{(i, k)}$ is the possibility of classifying $z_i$ to be of the same class as $z_{k}$, defined as below.
\begin{equation}
    \begin{aligned}
        q_{(i, k)} = &\frac{e^{z_i \cdot z_{k}/\tau}}{\sum_{a \in A(i)}^{}e^{ z_i \cdot z_a/\tau}}.
    \end{aligned}
    \label{eq:q}
\end{equation}
Eq.~\eqref{eq:grad_ind} reveals that for the positive samples, the gradient of $\mathcal{L}_{\left( i, p \right)}$ depends on the position of $z_k$. When $z_k$ appears in both numerator and denominator, \ie, as an effective positive, the gradient creates an attractive force to pull $z_k$ closer to $z_{i}$. Conversely, for the non-effective positives that appear only in the denominator, a repulsion force will push $z_{k}$ away from $z_i$, leading to conflicting gradients.

Next, we analyze the gradient from an instance perspective, and compute the gradients of the averaged pairwise losses from all the positive samples as expressed in Eq.~\eqref{eq:grad}.
\begin{equation}
    \begin{aligned}
        \frac{\partial \mathcal{L}_{i}}{\partial z_{k}}=\frac{1}{\tau} \times 
        \begin{cases} 
        -(\frac{1}{|P(i)|}-q_{(i, k)})z_i, &\text{\space if \space} k \in P(i); \\
        {q_{(i, k)}} {z_i}, &\text{\space otherwise}.
        \end{cases}
    \end{aligned}
    \label{eq:grad}
\end{equation}
We primarily focus on the gradients of positive samples.
\begin{equation}
    \begin{aligned}
        \left. \frac{\partial \mathcal{L}_i}{\partial z_{k}} \right|_{k \in P(i)}
        =&-\frac{z_i}{\tau}(\frac{1}{{|P(i)|}}- q_{(i, k)}).
    \end{aligned}
    \label{eq:grad_pos}
\end{equation}
Specifically, the sign of Eq.~\eqref{eq:grad_pos} determines the gradient direction towards positive sample $z_k$. We denote the second term as $\nabla$, where a positive $\nabla$ generates an attractive force and a negative $\nabla$ produces a repulsive one.
\begin{equation}
    \begin{aligned}
        \nabla=&\frac{1}{|P(i)|}- q_{(i, k)}. 
    \end{aligned}
    \label{eq:delta}
\end{equation}
In the beginning of training, the probability of $z_i$ and $z_k$ belonging to the same class is close to zero, making $\nabla > 0$. Hence $z_k$ will be pulled towards $z_i$, promoting the learning of a compact feature space for samples within the same class. 
As the model converges, class features collapse to the vertices of a simplex equiangular tight frame~\cite{liu2023inducing,yang2022inducing} and logits between inter-class features approach zero. In this situation $\nabla$ becomes:
\begin{equation}
    \begin{aligned}
        \nabla 
        =&\frac{1}{|P(i)|}- \frac{e^{z_i \cdot z_{k}/\tau}}{\sum_{p \in P(i)}^{}e^{z_i \cdot z_p/\tau}+\sum_{n \in A(i)\setminus P(i)}^{}e^{ z_i \cdot z_n /\tau}} \\
        \approx&\frac{1}{|P(i)|}- \frac{e^{z_i \cdot z_{k}/\tau}}{\sum_{p \in P(i)}^{}e^{ z_i \cdot z_p/\tau}} \\
        \approx&\frac{1}{|P(i)|}- \frac{e^{z_i \cdot z_{k}/\tau}}{|P(i)| e^{z_i \cdot z_{\bar{p}} /\tau}}
    \end{aligned}
    \label{eq:delta2}
\end{equation}
where $z_{\bar{p}}$ represents the expected feature center of all positive pairs in the current mini-batch.

\subsection{Problems of SCL in long-tailed recognition}
\label{sec:3.4}
\noindent\textbf{Conflicting gradients for easy positives}. As demonstrated in Eq.~\eqref{eq:delta}, when $z_k$ is distributed closer to sample $z_i$ compared with $z_{\bar{p}}$, \ie, an easy positive, $\nabla$ becomes negative. The generated conflicting gradient pushes $z_k$ away from its positive sample $z_i$, as depicted in Fig.~\ref{fig:scl_acl} (a).
However, it is unnecessary for our objective, \ie, to push $z_{k}$ and $z_{i}$ towards the class center. Instead, we could explicitly optimize $z_{i}$ and $z_{k}$ to be closer to their class centers.
On the other hand, easy samples typically contain representative semantic features that stabilize training and facilitate convergence. Thus, the repulsion of easy positives impedes the effective learning of robust and invariant features.

\noindent\textbf{Imbalanced attraction and repulsion gradients}. Due to the uneven distribution of positive pairs (\ie, $|P(i)|$) and negative pairs (\ie, batch-size$-1-|P(i)|$) across classes, SCL suffers from imbalanced gradients between attraction and repulsion terms. 
At the batch level, head classes contain more positive pairs to attract intra-class samples, yet fewer negative pairs to push samples away from other classes. This disparity results in strong intra-class compactness but potentially weak inter-class separation ability. In contrast, tail classes exhibit weak intra-class compactness but strong inter-class separation.


\section{Long-tailed recognition with multi-view}
\label{sec:4}
\noindent{\bf Multi-view training benefits long-tailed recognition.}
Contrastive learning constructs positive pairs with two augmented views. Then works~\cite{caron2020unsupervised, chen2020big,chen2021empirical} explore to make use of multiple views for pursuing good representations in self-supervised learning. In this paper, \textit{we observe that multi-view training can significantly enhance the performance of long-tailed recognition.} As shown in Fig.~\ref{fig:view}, with the number of views increasing from 1 to 4,  we achieve significant improvements of around 
5\% top-1 accuracy on ImageNet-LT with Balanced Softmax ~\cite{ren2020balanced}.
Multiple views can enhance the diversity of training data by generating varied representations of the same class. This helps the model learn more robust features, especially for the tail classes where original data is scarce.
Interestingly, we observe that \textit{contrastive learning's performance gains plateau when increasing views from 3 to 4}, 
which inspires us to delve deep into understanding the reasons behind it. 

\noindent{\bf Problems of multi-view training in SCL.}
In multi-view training, the number of positive pairs grows quadratically with class frequency as the number of views increases.
Denote the number of samples from the $j$-th class in the mini-batch as $n_j$.
For conventional two-view training, the total number of positive pairs from the $j$-th class is $2n_j(2n_j-1)$. This number increases to $mn_j(mn_j-1)$ when $m$ views are used. Denoting the increment in positive pairs as $T$, we then have $T \propto n_j^2$. 
Let $\beta$ represent the probability that sample $z_p$ is an easy positive compared with the expected class center $z_{\bar{p}}$ (\ie, $z_i \cdot z_p > z_i \cdot z_{\bar{p}}$). According to Eq.~\eqref{eq:delta}, the total number of conflicting gradients under two-view is $2n_j(2n_j-1) \beta$. In the multi-view scenario, the increment of conflicting gradients is also quadratic to the class frequency, \ie, $T \beta \propto n_j^2$. This demonstrates that multi-view training exacerbates the conflicting gradient problem by introducing more accessible positive pairs.
\section{Aligned contrastive learning}
\noindent{\bf Aligned Contrastive Loss.}
\label{sec:5}
Tackling the above drawbacks of SCL under the multi-view training setting, we propose a novel ACL loss with the following key designs:
\begin{itemize}
    \item ACL mitigates the conflict between effective and non-effective positive pairs by including only the effective pair in the denominator of the loss function. This modification promotes consistent attraction for all the positives. Meanwhile, we incorporate class centers into the contrastive process to explicitly encourage samples to be clustered. These class centers are dynamically updated in each batch using an exponential moving average, ensuring they remain representative of evolving class features.
    \item To achieve equilibrium between attraction and repulsion gradients, we propose re-weighting the negative pairs based on inverse class frequency. It ensures a balanced ratio of positive to negative pairs across different classes.
    Moreover, to balance positive pairs across classes, we modify the multi-view training strategy to be distribution-aware, assigning more views to underrepresented classes. Following ~\cite{caron2020unsupervised}, we utilize diverse scales for different views. 
\end{itemize}

Specifically, our ACL loss is formulated as:
\begin{equation}
    \begin{aligned}
         \mathcal{L}_{i}
        &=\frac{-1}{|P(i)|+1} \times \\
        &\sum_{p \in \{ {P(i),c} \} }\mathrm{log}\frac{e^{z_i\cdot z_p /\tau}}{e^{z_i\cdot z_p /\tau}+\sum\limits_{n \in N(i)}w_n e^{ z_i \cdot z_n /\tau}}
    \end{aligned}
    \label{eq:acl}
\end{equation}
where $c$ is the index for the class center, $N(i)$ is the set of all negatives containing samples and centers from other classes, and $w_n$ is the weight of each negative pair, which is inversely proportional to the class frequency.


\noindent{\bf The overall framework.}
\begin{figure}[t!]
    \centering 
    \includegraphics[width=0.85\linewidth]{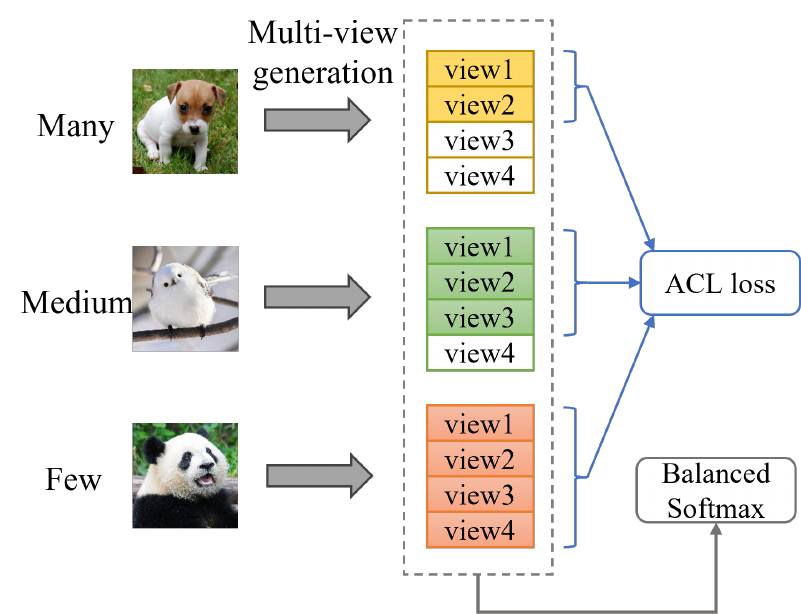} 
    \vspace{-1mm}
    \caption{Framework of the proposed ACL method. 
    \textcolor{myred}{Distribution-aware multi-views are selected for different sub-groups to compute the ACL loss. Concurrently, all views contribute to the Balanced Softmax loss calculation for robust classifier learning.}} 
    \label{fig:framework}
    \vspace{-2mm}
\end{figure}
An overview of the proposed framework is illustrated in Fig.~\ref{fig:framework}.
Distribution-aware multi-view is utilized in ACL computation to balance the contrastive pairs. Concurrently, all views contribute to the Balanced Softmax loss calculation, facilitating the learning of a robust and balanced classifier.
The overall loss is given below where $\alpha$ is the loss weight of ACL. 
\begin{equation}
    \begin{aligned}
        \mathcal{L} = \alpha \times \mathcal{L}_{acl} + \mathcal{L}_{bs}
    \end{aligned}
    \label{eq:loss}
\end{equation}

\noindent{\bf Analysis of ACL loss.}
We compute the gradient of ACL on the positive sample $z_p$ as shown below.
\begin{equation}
    \begin{aligned}
        \left. \frac{\partial \mathcal{L}_i}{\partial z_{p}} \right|_{p \in \{P(i),c\}}
        =& \frac{-1}{\tau (|P(i)|+1)} (1-q_{(i, k)})z_i\\
    \end{aligned}
    \label{eq:grad_acl}
\end{equation}
where $q_{(i,k)}$ is the aligned probability of $z_i$ and $z_k$ belonging to the same class with only the effective positive $z_k$ in the denominator:
\begin{equation}
    \begin{aligned}
        q_{(i, k)} = &\frac{e^{z_i \cdot z_{k}/\tau}}{e^{z_i\cdot z_k /\tau}+\sum\limits_{n \in N(i)}w_n e^{ z_i \cdot z_n /\tau}}
    \end{aligned}
    \label{eq:q_acl}
\end{equation}
Given that $1\geqslant q_{(i,k)} \geqslant 0$, Eq.~\eqref{eq:grad_acl} ensures consistent attracting gradients for all the positives, thereby eliminating the conflicting gradients present in SCL. 

\section{Experiments}
\label{sec:exp}
\subsection{Datasets}
\label{sec:6.1}

\noindent{\bf CIFAR-100-LT} is the long-tailed variant of CIFAR dataset. 
To quantify the degree of data imbalance, we adopted the imbalance factor (IF), defined as the ratio between the sample sizes of the most and least frequent classes, \ie, $IF=\frac{N_{max}}{N_{min}}$. Following~\cite{cui2021parametric, cui2023generalized, zhu2022balanced, du2024probabilistic}, we conducted experiments with IF of 10, 50, and 100. 

\noindent{\bf ImageNet-LT} is curated from the balanced ImageNet dataset by sampling under Pareto distribution with a power value of 6~\cite{liu2019large}. It comprises 115.8K images from 1,000 categories, with class sizes ranging from 5 to 1,280 samples. 

\noindent{\bf iNaturalist 2018} is a large-scale collection with an inherently imbalanced label distribution ~\cite{van2018inaturalist}. It comprises 437.5K images across 8,142 categories, exhibiting both long-tailed imbalance and fine-grained class distinctions.

\noindent\textcolor{myred}{{\bf Places-LT} is the long-tailed version of scene classification dataset~\cite{zhou2017places}, consisting of 184.5K images from 365 categories with class sizes ranging from 5 to 4,980.}

\subsection{Implementation details}
\label{sec:6.2}
For CIFAR-100-LT, we used ResNet-32 as the backbone architecture. Following~\cite{cui2023generalized, ren2020balanced}, we implemented AutoAugment~\cite{cubuk1805autoaugment} and CutOut~\cite{devries2017improved} for data augmentation. The initial learning rate was set to 0.07 with a linear warm-up for the first 10 epochs, followed by decay at epochs 160 and 180 with a step size of 0.1. We employed the SGD optimizer with a momentum of 0.9 and weight decay of 5e-4. The dimensions of MLP's hidden and output layer were 64 and 32, respectively. All other hyperparameters were kept consistent with the work~\cite{cui2023generalized}.

For ImageNet-LT, we employed ResNet-50~\cite{he2016deep}, ResNeXt-50-32x4d~\cite{xie2017aggregated}, and ResNeXt-101~\cite{xie2017aggregated} as backbone architectures, following previous works~\cite{cui2023generalized,kang2019decoupling}. We adopted a cosine classifier with normalized features and weight vectors. The MLP feature dimension was set to 1024. Models were trained for 90 epochs using SGD optimizer (momentum: 0.9, weight decay: 1e-3) with an initial learning rate of 0.05, decayed by a cosine scheduler. We trained the models with SGD and a batch size of 128. For the augmentation strategies in group-wise view, we referred to PaCo~\cite{cui2021parametric} and GPaCo~\cite{cui2023generalized} and adopted RandAug for the first view and RandAugStack for subsequent views.

For iNaturalist 2018, we trained models with ResNet-50 using the SGD optimizer with a momentum of 0.9. The models were trained with a total batch size of 128 on 4 GPUs. The initial learning rate was set to 0.04, with a 0.1 step-wise decay at epochs 120 and 160. 
\textcolor{myred}{For Places-LT, we used ResNet-152 pretrained on the full ImageNet-2012 dataset as the backbone. We strictly follow the settings of ~\cite{cui2023generalized} for fair comparison.}

For the settings of ACL, we set the loss weight $\alpha$ to 0.1 for CIFAR-LT and 0.5 for the rest datasets. 
We implemented distribution-aware multi-view training with 2, 3, and 4 views for many-shot ($>$100 samples per class), medium-shot (20$\sim$100 samples per class) and few-shot ($<$20 samples per class) categories, respectively. We referred to~\cite{caron2020unsupervised} and used varying scales for multiple views to reduce computational costs.

\subsection{Results}
\label{sec:6.3}
\begin{table}[t]
\centering
\scalebox{0.9}{
\setlength\tabcolsep{1.5pt}
\def\arraystretch{1.1}
\hspace*{-1em}
\begin{tabular}{p{3.3cm}<{\centering}p{1.8cm}<{\centering}p{1.8cm}<{\centering}p{1.8cm}<{\centering}}
\hline\hline
Method  & \multicolumn{3}{c}{CIFAR-100-LT}    \\ 
\hline 
IF & 100 & 50 & 10 \\
\hline
BBN~\cite{zhou2020bbn}        & 42.6  & 57.0   & 59.1    \\
Causal Model~\cite{tang2020long}     & 44.1  & 50.3   & 59.6    \\
LADE~\cite{hong2021disentangling}     & 45.4  & 50.5   & 61.7    \\
MiSLAS~\cite{zhong2021improving}     & 47.0  & 52.3   & 63.2     \\
Balanced Softmax~\cite{ren2020balanced}     & 50.8  & 54.2   & 63.0    \\
PaCo~\cite{cui2021parametric}     & 52.0  & 56.0   &  64.2     \\
BCL~\cite{zhu2022balanced}     & 51.9  & 56.6   & 64.9    \\
GPaCo~\cite{cui2023generalized}     & 52.3  & 56.4   & 65.4        \\
Ours-ACL    & \cellcolor{mygray}\textbf{52.6 (+0.3)}  & \cellcolor{mygray}\textbf{56.9 (+0.5)}    & \cellcolor{mygray}\textbf{66.4 (+1.0)}       \\
\hline \hline
\end{tabular}}
\vspace{-2mm}
\caption{Top-1 accuracy (\%) of ResNet-32 on CIFAR-100-LT.}
\label{tab:cifar}
\vspace{-3mm}
\end{table}

\noindent\textbf{Comparison on CIFAR-100-LT}. Table~\ref{tab:cifar} lists the comparison results between the proposed method and other existing works on CIFAR-100-LT. 
We observe that ACL is robust to imbalance factors and consistently outperforms previous long-tailed recognition methods.
Specifically, ACL surpasses the existing SOTA work GPaCo~\cite{cui2023generalized} by 0.3\%, 0.5\%, and 1.0\% under imbalance factors 100, 50, and 10 respectively, which testify the effectiveness of our method.

\begin{table}[t]
\centering
\scalebox{0.88}{
\setlength\tabcolsep{1.5pt}
\def\arraystretch{1.1}
\hspace*{-1em}
\begin{tabular}{p{3.45cm}<{\centering}p{1.7cm}<{\centering}p{1.9cm}<{\centering}p{1.95cm}<{\centering}}
\hline\hline
Method   & ResNet-50 & ResNeXt-50 & ResNeXt-101  \\
\hline
Cross Entropy~\cite{zhou2020bbn}        & 41.6  & 44.4   & 44.8    \\
Decouple~\cite{kang2019decoupling}     & 46.7  & 49.4   & 49.6    \\
Causal Model~\cite{tang2020long}     & 51.7  & 51.8   & 53.3    \\
DisAlign~\cite{zhang2021distribution}     & 52.9  & 53.4   & -      \\
BCL~\cite{zhu2022balanced}     & 56.0  & 56.7  & -   \\
DSCL~\cite{xuan2024decoupled}     & 57.7  & 58.7   & -    \\
ProCo~\cite{du2024probabilistic}       & 57.3  & 58.0   & -    \\
Balanced Softmax*~\cite{ren2020balanced}     & 55.0  & 56.2   & 58.0    \\
PaCo*~\cite{cui2021parametric}     & 57.0  & 58.2  & 60.0      \\
GPaCo*~\cite{cui2023generalized}     & 58.5  & 58.9   & 60.8       \\
Ours-ACL    & \cellcolor{mygray}\textbf{59.7 (+1.2)}  & \cellcolor{mygray}\textbf{61.1 (+2.2)}    & \cellcolor{mygray}\textbf{61.9 (+1.1)}       \\
\hline \hline
\end{tabular}}
\vspace{-2mm}
\caption{Top-1 accuracy (\%) on ImageNet-LT for different backbone architectures. (“*”: models trained under 400 epochs)}
\label{tab:imagenet_all}
\end{table}

\begin{table}[t]
\centering
\scalebox{0.9}{
\setlength\tabcolsep{1.5pt}
\def\arraystretch{1.1}
\hspace*{-1em}
\begin{tabular}{p{3.3cm}<{\centering}p{1.3cm}<{\centering}p{1.3cm}<{\centering}p{1.3cm}<{\centering}p{1.3cm}<{\centering}}
\hline\hline
Method   & Many & Medium & Few & All \\
\hline
Balanced Softmax~\cite{ren2020balanced}     & 62.2  & 48.8   & 29.8      & 51.4   \\
LADE~\cite{hong2021disentangling}     & 62.3  & 49.3   & 31.2      & 51.9   \\
Causal Model~\cite{tang2020long}     & 62.7  & 48.8   & 31.6      & 51.8   \\
DisAlign~\cite{zhang2021distribution}     & 62.7  & 52.1   & 31.4      & 53.4   \\
BCL~\cite{zhu2022balanced}     & 67.2  & 53.9   & 36.5      & 56.7   \\
GML~\cite{suh2023long}     & 68.7  & 55.7   & 38.6      & 58.3   \\
PaCo*~\cite{cui2021parametric}     & 68.0  & 70.0   & 56.4    & 58.2   \\
GPaCo*~\cite{cui2023generalized}     & 67.9  & 57.1   & 40.1      & 58.9   \\
Ours-ACL      & \cellcolor{mygray}\textbf{\makecell[c] {70.7\\(+2.8)}}  & \cellcolor{mygray}\textbf{\makecell[c]{59.1\\(+2.0)}}    & \cellcolor{mygray}\textbf{\makecell[c]{40.6\\(+0.5)}}      & \cellcolor{mygray}\textbf{\makecell[c]{61.1\\(+2.2)}}    \\
\hline \hline
\end{tabular}}
\vspace{-2mm}
\caption{Top-1 accuracy (\%) of ResNext-50 on ImageNet-LT. (“*”: models trained under 400 epochs)}
\label{tab:imagenetX50}
\vspace{-3mm}
\end{table}
\noindent\textbf{Comparison on ImageNet-LT}. We conducted extensive experiments on ImageNet-LT with different backbone architectures, and the results are presented in Table~\ref{tab:imagenet_all}. 
Our ACL method consistently outperforms existing approaches across different backbones, achieving superior overall performance with significant margins. Notably, compared to GPaCo~\cite{cui2023generalized}, another method based on contrastive learning, ACL improves the overall top-1 accuracy by more than 1\% across all tested architectures.

In addition, we report the group-wise accuracy on each category in Table~\ref{tab:imagenetX50}. 
ACL significantly outperforms the baseline Balanced Softmax method, validating the effectiveness of contrastive learning in boosting overall performance. 
Compared to other contrastive learning-based approaches like BCL~\cite{zhu2022balanced}, DSCL~\cite{xuan2024decoupled}, PaCo~\cite{cui2021parametric}, and GPaCo~\cite{cui2023generalized}, ACL achieves superior accuracy across all categories. Specifically, ACL surpasses the current SOTA method GPaCo with remarkable improvements of 2.5\%, 2.0\%, and 0.5\% in many-shot, medium-shot, and few-shot respectively, setting a new benchmark with 61.1\% overall accuracy. 
These results suggest that ACL effectively eliminates conflict gradients and balances gradient contributions across different pairs and classes, thereby fully leveraging contrastive learning to develop robust features.

\begin{table}[t]
\centering
\scalebox{0.9}{
\setlength\tabcolsep{1.5pt}
\def\arraystretch{1.1}
\hspace*{-1em}
\begin{tabular}{p{3.5cm}<{\centering}p{2.0cm}<{\centering}p{2.0cm}<{\centering}}
\hline\hline

Method  & \multicolumn{2}{c}{Top-1 accuracy}    \\ 
\hline 
Dataset & iNaturalist & \textcolor{myred}{Places-LT} \\
\hline
Cross Entropy        & 61.7 & 30.2 \\
KCL~\cite{kang2020exploring}        & 68.6 & -  \\
BBN~\cite{zhou2020bbn}     & 69.6   & - \\
MiSLAS~\cite{zhong2021improving}     & 71.6   & 40.4 \\
Balanced Softmax~\cite{ren2020balanced}     & 71.8   & 38.6\\
BCL~\cite{zhu2022balanced}     & 71.8    & - \\
PaCo~\cite{cui2021parametric}     & 73.2   & 41.2 \\
ProCo~\cite{du2024probabilistic}       & 73.5  & - \\
GML~\cite{suh2023long}   & 74.5 & - \\
GPaCo~\cite{cui2023generalized}     & 75.4       & 41.7 \\
Ours-ACL    & \cellcolor{mygray}\textbf{75.6 (+0.2)}     & \cellcolor{mygray}\textbf{42.4 (+0.7)} \\
\hline \hline
\end{tabular}}
\vspace{-2mm}
\caption{Top-1 accuracy (\%) on iNaturalist 2018 and \textcolor{myred}{Places-LT}.}
\label{tab:inat}
\end{table}

\noindent\textbf{Comparison on iNaturalist 2018 and \textcolor{myred}{Places-LT}}. 
Table~\ref{tab:inat} shows the experimental results on iNaturalist 2018 and Places-LT.
On iNaturalist 2018, our method consistently outperforms recent SOTA approaches like BCL~\cite{zhu2022balanced} and GML~\cite{suh2023long}, achieving competitive performance with GPaCo~\cite{cui2023generalized}. 
\textcolor{myred}{On Places-LT, the top-1 accuracy of ACL greatly surpasses GPaco by 0.7\%, validating the effectiveness of our method.}


\subsection{Ablation study}

\begin{table}[t]
\centering
\scalebox{0.9}{
\setlength\tabcolsep{1.5pt}
\def\arraystretch{1.1}
\hspace*{-1em}
\begin{tabular}{p{1.8cm}<{\centering}p{1.4cm}<{\centering}p{1.4cm}<{\centering}p{1.4cm}<{\centering}p{1.3cm}<{\centering}}
\hline\hline
\# of views   & Many & Medium & Few & All \\
\hline
1     & 63.9  & 52.2   & 35.1      & 54.4   \\
2     & 67.9  & 55.7   & 37.7      & 58.0   \\
3     & 69.5  & 57.1   & 38.6      & 59.4   \\
4     & 70.6  & 58.3   & 39.5      & 60.4   \\
5     & 71.0  & 58.5   & 39.7      & 60.7   \\
\hline \hline
\end{tabular}}
\vspace{-2mm}
\caption{Performance of baseline models on ImageNet-LT with various views (ResNeXt-50 backbone).}
\label{tab:view}
\vspace{-2mm}
\end{table}
\noindent\textbf{Number of views in multi-view training}.
We built our ACL based on the baseline model of Balanced Softmax loss as described in Fig.~\ref{fig:framework}. To determine the optimal number of views for multi-view training, we trained baseline models with varying numbers of views, as shown in Table~\ref{tab:view}. As the number of views increases, performance across all categories and top-1 accuracy improve consistently. Since 4 views yielded comparable results to 5 views, we chose 4 views for subsequent experiments to balance performance and computational efficiency.

\begin{table}[t]
\centering
\scalebox{0.9}{
\setlength\tabcolsep{1.5pt}
\def\arraystretch{1.1}
\hspace*{-1em}
\begin{tabular}{p{3.6cm}<{\centering}p{1.2cm}<{\centering}p{1.2cm}<{\centering}p{1.2cm}<{\centering}p{1.1cm}<{\centering}}
\hline\hline
Method   & Many & Medium & Few & All \\
\hline
Baseline     & 70.6  & 58.3   & 39.5      & 60.4   \\
+SCL*     & 69.9  & 57.6   & 40.1      & 59.9   \\
+SCL$^\dagger$     & 70.1  & 58.3   & 40.6      & 60.4   \\
+ACL (Ours)     & \cellcolor{mygray}\textbf{70.7}  & \cellcolor{mygray}\textbf{59.1}    & \cellcolor{mygray}\textbf{40.6}      & \cellcolor{mygray}\textbf{61.1}    \\
\hline \hline
\end{tabular}}
\vspace{-2mm}
\caption{Performance comparison between SCL and ACL under multi-view training. Results are from ResNeXt-50 on ImageNet-LT. (“*”: uniform multi-view. “$^\dagger$”: distribution-aware multi-view.)}
\label{tab:scl_acl}
\vspace{-4mm}
\end{table}
\noindent\textbf{Comparison with SCL}.
To evaluate ACL's efficacy in mitigating conflicting gradients, we compare it with SCL under the multi-view training setting. The baseline is constructed with 4 views using Balanced Softmax. We implemented SCL with both uniform multi-view across all classes and distribution-aware views as used in ACL. Table~\ref{tab:scl_acl} shows that SCL leads to performance degradation compared to the baseline model, particularly in many-shot category. As discussed in Section~\ref{sec:4}, this results from conflicting gradients in classes with numerous positive pairs. While distribution-aware multi-view partially mitigates this issue by rebalancing the contrastive pairs distribution, it fails to surpass the baseline. This indicates that rebalancing alone is insufficient to resolve gradient conflicts in SCL. Our proposed ACL effectively eliminates the conflicting gradients and promotes consistent attraction among all the positives and class center, yielding improvements across all categories.

\begin{figure*}
  \centering
  \begin{subfigure}{0.33\linewidth}
    { \includegraphics[width=0.99\linewidth]{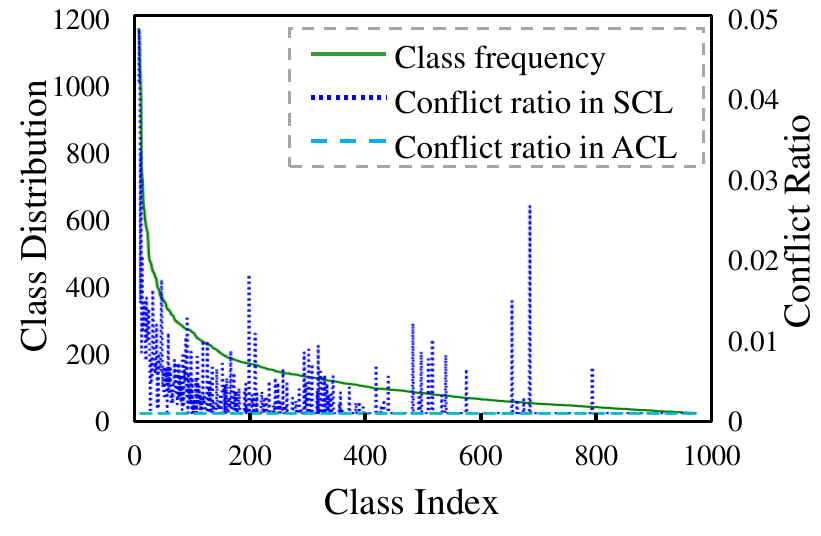}}
    \caption{}
    \label{fig:grad}
  \end{subfigure}
  \hfill
  \begin{subfigure}{0.33\linewidth}
    { \includegraphics[width=0.99\linewidth]{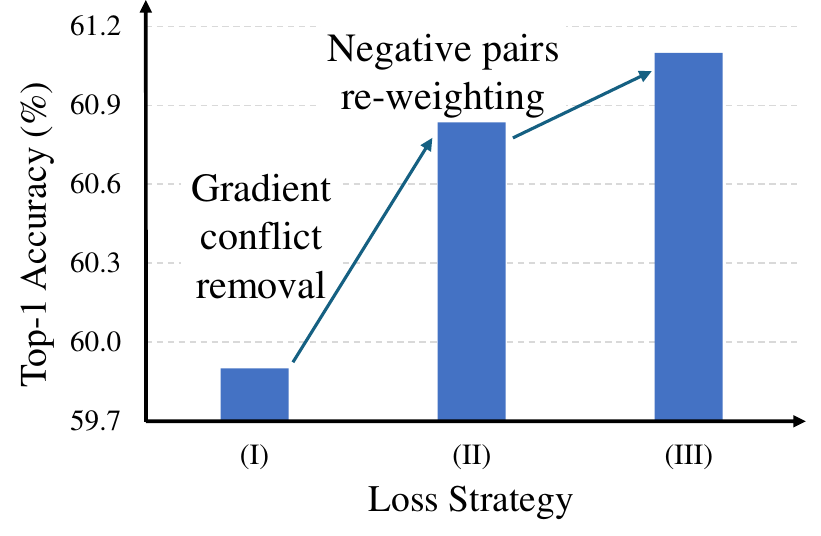}}
    \caption{}
    \label{fig:ablation}
  \end{subfigure}
  \hfill
  \begin{subfigure}{0.33\linewidth}
    { \includegraphics[width=0.99\linewidth]{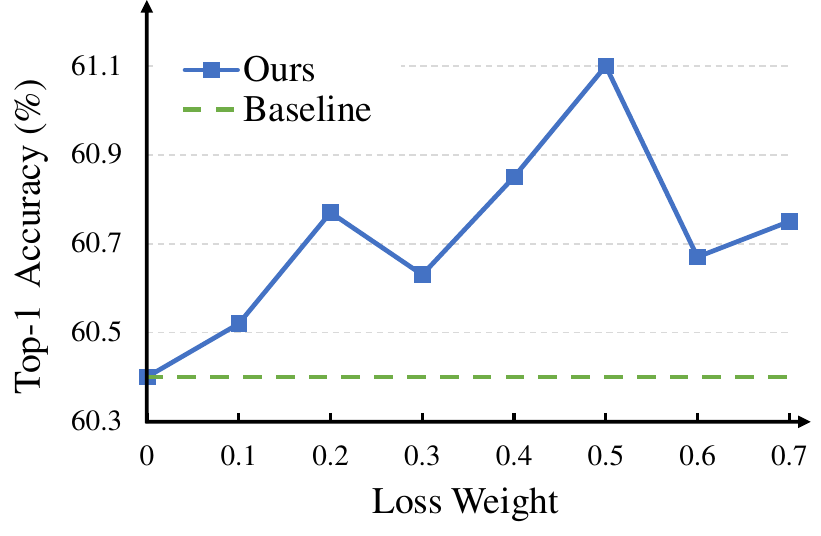}} 
    \caption{}
    \label{fig:alpha}
  \end{subfigure}
  \vspace{-8mm}
  \caption{(a) Relationship between the conflict ratio of gradients and the class frequency distribution on the ImageNet-LT dataset;
        (b) Top-1 accuracy (\%) of models trained with different loss settings. (I) SCL loss with gradient conflict. (II) ACL loss with consistent gradient. (III) ACL loss with consistent gradient and negative pairs re-weighting;
        (c) Top-1 accuracy (\%) on ImageNet-LT dataset with different loss weight $\alpha$ (ResNeXt-50 backbone).}
  \label{fig:short}
  \vspace{-3mm}
\end{figure*}

\noindent\textbf{Gradient monitoring}. 
In addition to the theoretical analysis in Section~\ref{sec:3.3}, we monitored the ratio of conflicting gradients in each class during actual training. 
Fig.~\ref{fig:grad} illustrates the relationship between the conflict occurrence and the class distribution. We observe a positive correlation where more frequent classes experience a higher incidence of conflicts. This aligns with our previous analysis, which suggests that more positive pairs with multi-views would exacerbate gradient conflicts. 
The proposed ACL effectively eliminates gradient conflicts and enhances overall performance. 
Table~\ref{tab:scl_acl} indicates that ACL yields greater performance improvements over SCL in many-shot and medium-shot classes compared to few-shot classes, where gradient conflicts are less severe. These results validate ACL's effectiveness in addressing gradient conflict issues.


\noindent\textbf{Effectiveness of each strategy}.
To analyze the effectiveness of each strategy proposed in Section~\ref{sec:5}, we conducted an ablation study with different loss settings. As illustrated in Fig.~\ref{fig:ablation}, SCL in  model (I) suffers from conflicting gradients limiting its efficacy in learning robust representations.
By eliminating this inherent conflict, model (II) enjoys consistent attraction among positives and class centers, yielding significant improvement. 
Further application of re-weighting to negative pairs in our ACL results in additional performance gains as shown by model (III).

\begin{table}[t]
\centering
\scalebox{0.9}{
\setlength\tabcolsep{1.5pt}
\def\arraystretch{1.1}
\hspace*{-1em}
\begin{tabular}{p{2.2cm}<{\centering}p{1.5cm}<{\centering}p{1.5cm}<{\centering}p{1.5cm}<{\centering}p{1.5cm}<{\centering}}
\hline\hline
Loss weight   & Many & Medium & Few & All \\
\hline
0.2     & 70.9  & 58.3   & 39.6      & 60.7   \\
0.5     & 70.7  & 59.1   & 40.6      & 61.1   \\
0.8     & 70.4  & 58.7   & 40.8      & 60.8   \\
\hline \hline
\end{tabular}}
\vspace{-3mm}
\caption{\textcolor{myred}{Results on ImageNet-LT with different loss weights.}}
\label{tab:tradeoff}
\vspace{-0.1in}
\end{table}

\noindent\textbf{Effect of loss weight}.
The impact of loss weight $\alpha$ is shown in Fig.~\ref{fig:alpha}.
While the baseline Balanced Softmax loss clusters samples around class centers without explicit similarity constraints, contrastive learning directly enforces similarity between same-class samples in latent space. The results show that increasing ACL strength initially improves top-1 accuracy before causing degradation, with $\alpha=0.5$ emerging as the optimal value in our experiments.

\textcolor{myred}{Moreover, the loss weight can also trade-off between head and tail class accuracy as shown in Table~\ref{tab:tradeoff}. A larger $\alpha$ leads to higher performance of tail classes at some cost of head class accuracy.}



\begin{figure}[t!]
    \centering 
    \includegraphics[width=0.9\linewidth]{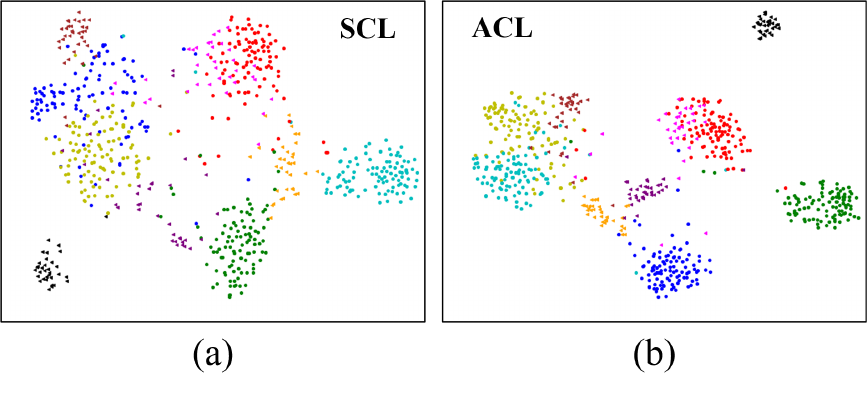} 
    \vspace{-5mm}
    \caption{Feature visualization of CIFAR-100-LT validation data with IF of 10. (a) and (b) are the t-SNE results from SCL and our ACL models respectively (best viewed in color).} 
    \label{fig:tsne}
    \vspace{-3mm}
\end{figure}
\noindent\textbf{Visualization of learned features}.
We visualized the features of SCL and ACL models under multi-view training on CIFAR-100-LT using t-SNE. We randomly selected five classes each from the many-shot and few-shot category for clarity, as illustrated in Fig.~\ref{fig:tsne}. Different colors represent distinct classes, with more populated colors indicating many-shot classes and fewer colors denoting few-shot classes.
Our analysis reveals that ACL reduces the overlapping between different classes, and leads to more compact and separable representations.

\begin{table}[t]
\centering
\scalebox{0.9}{
\setlength\tabcolsep{1.5pt}
\def\arraystretch{1.1}
\hspace*{-1em}
\begin{tabular}{p{2.2cm}<{\centering}p{1.6cm}<{\centering}p{1.6cm}<{\centering}p{1.6cm}<{\centering}p{1.6cm}<{\centering}}
\hline\hline
Datasets   & \makecell[c]{CIFAR-\\100-LT} & \makecell[c]{ImageNet-\\LT} & \makecell[c]{iNaturalist\\2018} & \makecell[c]{Places-\\LT} \\
\hline
GPaCo*                    &65.4   &58.9    &75.4   &41.7    \\
\hline
\makecell[c]{Multiview\\(Baseline)}    &65.6   &60.4    &74.6   &41.2    \\
ACL (Ours)                 &66.4(\textbf{+0.8})   &61.1(\textbf{+0.7})    &75.6(\textbf{+1.0})       &42.4(\textbf{+1.2})    \\
\hline \hline
\end{tabular}}
\vspace{-3mm}
\caption{\textcolor{myred}{ACL significantly outperforms the multi-view training baseline. (“*”: models trained under 400 epochs)}}
\label{tab:acl_improvements}
\vspace{-3mm}
\end{table}
\noindent\textbf{\textcolor{myred}{ACL significantly outperforms multi-view training baseline.}}
\textcolor{myred}{It is worth noting that PaCo/GPaCo models are trained in 400 epochs. We train ACL models with fewer epochs, \eg, 200 epochs on iNaturalist 2018.
As shown in Table~\ref{tab:acl_improvements}, we consistently achieve significant improvements over the multi-view training baseline (multi-view and multi-task learning with SCL) on CIFAR-100-LT, ImageNet-LT, iNaturalist 2018, and Places-LT, surpassing the baseline models by 0.8\%, 0.7\%, 1.0\%, and 1.2\% individually. ACL with a longer training scheme can potentially further promote model performance.}

\noindent\textbf{\textcolor{myred}{Discussion on foundational vision-language models.}}
\textcolor{myred}{Vision-language foundation models primarily employ contrastive loss to align images with corresponding text in the embedding space without explicit class labels. While our proposed ACL for supervised learning is not directly applicable to the pretaining of foundation models, it offers potential improvements in robust representation learning when transferring pre-trained models to downstream tasks. Further details are provided in the appendix.}

\section{Conclusion}
\label{sec:conclusion}
In this work, we identify the conflicting gradients in conventional supervised contrastive loss (SCL) which causes performance degradation under the multi-view training setting. We propose a novel aligned contrastive loss (ACL) for long-tailed recognition. It eliminates the conflicts and promotes consistent attraction gradients among all the positive pairs and class centers. Furthermore, ACL achieves equilibrium between attraction and repulsion gradients. Experiments conducted across various benchmarks demonstrate that our method establishes a new SOTA in long-tailed recognition. 


{
    \small
    \bibliographystyle{ieeenat_fullname}
    \bibliography{main}

\begin{thebibliography}{51}
\providecommand{\natexlab}[1]{#1}
\providecommand{\url}[1]{\texttt{#1}}
\expandafter\ifx\csname urlstyle\endcsname\relax
  \providecommand{\doi}[1]{doi: #1}\else
  \providecommand{\doi}{doi: \begingroup \urlstyle{rm}\Url}\fi

\bibitem[Buda et~al.(2018)Buda, Maki, and Mazurowski]{buda2018systematic}
Mateusz Buda, Atsuto Maki, and Maciej~A Mazurowski.
\newblock A systematic study of the class imbalance problem in convolutional neural networks.
\newblock \emph{Neural networks}, 106:\penalty0 249--259, 2018.

\bibitem[Byrd and Lipton(2019)]{byrd2019effect}
Jonathon Byrd and Zachary Lipton.
\newblock What is the effect of importance weighting in deep learning?
\newblock In \emph{International conference on machine learning}, pages 872--881. PMLR, 2019.

\bibitem[Caron et~al.(2020)Caron, Misra, Mairal, Goyal, Bojanowski, and Joulin]{caron2020unsupervised}
Mathilde Caron, Ishan Misra, Julien Mairal, Priya Goyal, Piotr Bojanowski, and Armand Joulin.
\newblock Unsupervised learning of visual features by contrasting cluster assignments.
\newblock \emph{Advances in neural information processing systems}, 33:\penalty0 9912--9924, 2020.

\bibitem[Chen et~al.(2020{\natexlab{a}})Chen, Kornblith, Norouzi, and Hinton]{chen2020simple}
Ting Chen, Simon Kornblith, Mohammad Norouzi, and Geoffrey Hinton.
\newblock A simple framework for contrastive learning of visual representations.
\newblock In \emph{International conference on machine learning}, pages 1597--1607. PMLR, 2020{\natexlab{a}}.

\bibitem[Chen et~al.(2020{\natexlab{b}})Chen, Kornblith, Swersky, Norouzi, and Hinton]{chen2020big}
Ting Chen, Simon Kornblith, Kevin Swersky, Mohammad Norouzi, and Geoffrey~E Hinton.
\newblock Big self-supervised models are strong semi-supervised learners.
\newblock \emph{Advances in neural information processing systems}, 33:\penalty0 22243--22255, 2020{\natexlab{b}}.

\bibitem[Chen and He(2021)]{chen2021exploring}
Xinlei Chen and Kaiming He.
\newblock Exploring simple siamese representation learning.
\newblock In \emph{Proceedings of the IEEE/CVF conference on computer vision and pattern recognition}, pages 15750--15758, 2021.

\bibitem[Chen et~al.(2021)Chen, Xie, and He]{chen2021empirical}
Xinlei Chen, Saining Xie, and Kaiming He.
\newblock An empirical study of training self-supervised vision transformers.
\newblock In \emph{Proceedings of the IEEE/CVF international conference on computer vision}, pages 9640--9649, 2021.

\bibitem[Cubuk et~al.(1805)Cubuk, Zoph, Mane, Vasudevan, and Le]{cubuk1805autoaugment}
Ekin~D Cubuk, Barret Zoph, Dandelion Mane, Vijay Vasudevan, and Quoc~V Le.
\newblock Autoaugment: Learning augmentation policies from data. arxiv 2018.
\newblock \emph{arXiv preprint arXiv:1805.09501}, 2, 1805.

\bibitem[Cui et~al.(2021)Cui, Zhong, Liu, Yu, and Jia]{cui2021parametric}
Jiequan Cui, Zhisheng Zhong, Shu Liu, Bei Yu, and Jiaya Jia.
\newblock Parametric contrastive learning.
\newblock In \emph{Proceedings of the IEEE/CVF international conference on computer vision}, pages 715--724, 2021.

\bibitem[Cui et~al.(2023)Cui, Zhong, Tian, Liu, Yu, and Jia]{cui2023generalized}
Jiequan Cui, Zhisheng Zhong, Zhuotao Tian, Shu Liu, Bei Yu, and Jiaya Jia.
\newblock Generalized parametric contrastive learning.
\newblock \emph{IEEE Transactions on Pattern Analysis and Machine Intelligence}, 2023.

\bibitem[Cui et~al.(2019)Cui, Jia, Lin, Song, and Belongie]{cui2019class}
Yin Cui, Menglin Jia, Tsung-Yi Lin, Yang Song, and Serge Belongie.
\newblock Class-balanced loss based on effective number of samples.
\newblock In \emph{Proceedings of the IEEE/CVF conference on computer vision and pattern recognition}, pages 9268--9277, 2019.

\bibitem[DeVries and Taylor(2017)]{devries2017improved}
Terrance DeVries and Graham~W Taylor.
\newblock Improved regularization of convolutional neural networks with cutout.
\newblock \emph{arXiv preprint arXiv:1708.04552}, 2017.

\bibitem[Drummond et~al.(2003)Drummond, Holte, et~al.]{drummond2003c4}
Chris Drummond, Robert~C Holte, et~al.
\newblock C4. 5, class imbalance, and cost sensitivity: why under-sampling beats over-sampling.
\newblock In \emph{Workshop on learning from imbalanced datasets II}, pages 1--8, 2003.

\bibitem[Du et~al.(2024)Du, Wang, Song, and Huang]{du2024probabilistic}
Chaoqun Du, Yulin Wang, Shiji Song, and Gao Huang.
\newblock Probabilistic contrastive learning for long-tailed visual recognition.
\newblock \emph{IEEE Transactions on Pattern Analysis and Machine Intelligence}, 2024.

\bibitem[Fort et~al.(2021)Fort, Brock, Pascanu, De, and Smith]{fort2021drawing}
Stanislav Fort, Andrew Brock, Razvan Pascanu, Soham De, and Samuel~L Smith.
\newblock Drawing multiple augmentation samples per image during training efficiently decreases test error.
\newblock \emph{arXiv preprint arXiv:2105.13343}, 2021.

\bibitem[Grill et~al.(2020)Grill, Strub, Altch{\'e}, Tallec, Richemond, Buchatskaya, Doersch, Avila~Pires, Guo, Gheshlaghi~Azar, et~al.]{grill2020bootstrap}
Jean-Bastien Grill, Florian Strub, Florent Altch{\'e}, Corentin Tallec, Pierre Richemond, Elena Buchatskaya, Carl Doersch, Bernardo Avila~Pires, Zhaohan Guo, Mohammad Gheshlaghi~Azar, et~al.
\newblock Bootstrap your own latent-a new approach to self-supervised learning.
\newblock \emph{Advances in neural information processing systems}, 33:\penalty0 21271--21284, 2020.

\bibitem[He et~al.(2016)He, Zhang, Ren, and Sun]{he2016deep}
Kaiming He, Xiangyu Zhang, Shaoqing Ren, and Jian Sun.
\newblock Deep residual learning for image recognition.
\newblock In \emph{Proceedings of the IEEE conference on computer vision and pattern recognition}, pages 770--778, 2016.

\bibitem[He et~al.(2020)He, Fan, Wu, Xie, and Girshick]{he2020momentum}
Kaiming He, Haoqi Fan, Yuxin Wu, Saining Xie, and Ross Girshick.
\newblock Momentum contrast for unsupervised visual representation learning.
\newblock In \emph{Proceedings of the IEEE/CVF conference on computer vision and pattern recognition}, pages 9729--9738, 2020.

\bibitem[Hong et~al.(2021)Hong, Han, Choi, Seo, Kim, and Chang]{hong2021disentangling}
Youngkyu Hong, Seungju Han, Kwanghee Choi, Seokjun Seo, Beomsu Kim, and Buru Chang.
\newblock Disentangling label distribution for long-tailed visual recognition.
\newblock In \emph{Proceedings of the IEEE/CVF conference on computer vision and pattern recognition}, pages 6626--6636, 2021.

\bibitem[Hou et~al.(2023)Hou, Zhang, Wang, and Zhou]{hou2023subclass}
Chengkai Hou, Jieyu Zhang, Haonan Wang, and Tianyi Zhou.
\newblock Subclass-balancing contrastive learning for long-tailed recognition.
\newblock In \emph{Proceedings of the IEEE/CVF International Conference on Computer Vision}, pages 5395--5407, 2023.

\bibitem[Huang et~al.(2016)Huang, Li, Loy, and Tang]{huang2016learning}
Chen Huang, Yining Li, Chen~Change Loy, and Xiaoou Tang.
\newblock Learning deep representation for imbalanced classification.
\newblock In \emph{Proceedings of the IEEE conference on computer vision and pattern recognition}, pages 5375--5384, 2016.

\bibitem[Kang et~al.(2019)Kang, Xie, Rohrbach, Yan, Gordo, Feng, and Kalantidis]{kang2019decoupling}
Bingyi Kang, Saining Xie, Marcus Rohrbach, Zhicheng Yan, Albert Gordo, Jiashi Feng, and Yannis Kalantidis.
\newblock Decoupling representation and classifier for long-tailed recognition.
\newblock \emph{arXiv preprint arXiv:1910.09217}, 2019.

\bibitem[Kang et~al.(2020)Kang, Li, Xie, Yuan, and Feng]{kang2020exploring}
Bingyi Kang, Yu Li, Sa Xie, Zehuan Yuan, and Jiashi Feng.
\newblock Exploring balanced feature spaces for representation learning.
\newblock In \emph{International Conference on Learning Representations}, 2020.

\bibitem[Khan et~al.(2017)Khan, Hayat, Bennamoun, Sohel, and Togneri]{khan2017cost}
Salman~H Khan, Munawar Hayat, Mohammed Bennamoun, Ferdous~A Sohel, and Roberto Togneri.
\newblock Cost-sensitive learning of deep feature representations from imbalanced data.
\newblock \emph{IEEE transactions on neural networks and learning systems}, 29\penalty0 (8):\penalty0 3573--3587, 2017.

\bibitem[Khosla et~al.(2020)Khosla, Teterwak, Wang, Sarna, Tian, Isola, Maschinot, Liu, and Krishnan]{khosla2020supervised}
Prannay Khosla, Piotr Teterwak, Chen Wang, Aaron Sarna, Yonglong Tian, Phillip Isola, Aaron Maschinot, Ce Liu, and Dilip Krishnan.
\newblock Supervised contrastive learning.
\newblock \emph{Advances in neural information processing systems}, 33:\penalty0 18661--18673, 2020.

\bibitem[Li et~al.(2022)Li, Cao, Yuan, Fan, Yang, Feris, Indyk, and Katabi]{li2022targeted}
Tianhong Li, Peng Cao, Yuan Yuan, Lijie Fan, Yuzhe Yang, Rogerio~S Feris, Piotr Indyk, and Dina Katabi.
\newblock Targeted supervised contrastive learning for long-tailed recognition.
\newblock In \emph{Proceedings of the IEEE/CVF Conference on Computer Vision and Pattern Recognition}, pages 6918--6928, 2022.

\bibitem[Liu et~al.(2023)Liu, Zhang, Hu, Cao, Yao, and Pan]{liu2023inducing}
Xuantong Liu, Jianfeng Zhang, Tianyang Hu, He Cao, Yuan Yao, and Lujia Pan.
\newblock Inducing neural collapse in deep long-tailed learning.
\newblock In \emph{International Conference on Artificial Intelligence and Statistics}, pages 11534--11544. PMLR, 2023.

\bibitem[Liu et~al.(2019)Liu, Miao, Zhan, Wang, Gong, and Yu]{liu2019large}
Ziwei Liu, Zhongqi Miao, Xiaohang Zhan, Jiayun Wang, Boqing Gong, and Stella~X Yu.
\newblock Large-scale long-tailed recognition in an open world.
\newblock In \emph{Proceedings of the IEEE/CVF conference on computer vision and pattern recognition}, pages 2537--2546, 2019.

\bibitem[Menon et~al.(2020)Menon, Jayasumana, Rawat, Jain, Veit, and Kumar]{menon2020long}
Aditya~Krishna Menon, Sadeep Jayasumana, Ankit~Singh Rawat, Himanshu Jain, Andreas Veit, and Sanjiv Kumar.
\newblock Long-tail learning via logit adjustment.
\newblock \emph{arXiv preprint arXiv:2007.07314}, 2020.

\bibitem[Nam et~al.(2023)Nam, Jang, and Lee]{nam2023decoupled}
Giung Nam, Sunguk Jang, and Juho Lee.
\newblock Decoupled training for long-tailed classification with stochastic representations.
\newblock \emph{arXiv preprint arXiv:2304.09426}, 2023.

\bibitem[Oord et~al.(2018)Oord, Li, and Vinyals]{oord2018representation}
Aaron van~den Oord, Yazhe Li, and Oriol Vinyals.
\newblock Representation learning with contrastive predictive coding.
\newblock \emph{arXiv preprint arXiv:1807.03748}, 2018.

\bibitem[Pouyanfar et~al.(2018)Pouyanfar, Tao, Mohan, Tian, Kaseb, Gauen, Dailey, Aghajanzadeh, Lu, Chen, et~al.]{pouyanfar2018dynamic}
Samira Pouyanfar, Yudong Tao, Anup Mohan, Haiman Tian, Ahmed~S Kaseb, Kent Gauen, Ryan Dailey, Sarah Aghajanzadeh, Yung-Hsiang Lu, Shu-Ching Chen, et~al.
\newblock Dynamic sampling in convolutional neural networks for imbalanced data classification.
\newblock In \emph{2018 IEEE conference on multimedia information processing and retrieval (MIPR)}, pages 112--117. IEEE, 2018.

\bibitem[Ren et~al.(2020)Ren, Yu, Ma, Zhao, Yi, et~al.]{ren2020balanced}
Jiawei Ren, Cunjun Yu, Xiao Ma, Haiyu Zhao, Shuai Yi, et~al.
\newblock Balanced meta-softmax for long-tailed visual recognition.
\newblock \emph{Advances in neural information processing systems}, 33:\penalty0 4175--4186, 2020.

\bibitem[Shen et~al.(2016)Shen, Lin, and Huang]{shen2016relay}
Li Shen, Zhouchen Lin, and Qingming Huang.
\newblock Relay backpropagation for effective learning of deep convolutional neural networks.
\newblock In \emph{Computer Vision--ECCV 2016: 14th European Conference, Amsterdam, The Netherlands, October 11--14, 2016, Proceedings, Part VII 14}, pages 467--482. Springer, 2016.

\bibitem[Shu et~al.(2019)Shu, Xie, Yi, Zhao, Zhou, Xu, and Meng]{shu2019meta}
Jun Shu, Qi Xie, Lixuan Yi, Qian Zhao, Sanping Zhou, Zongben Xu, and Deyu Meng.
\newblock Meta-weight-net: Learning an explicit mapping for sample weighting.
\newblock \emph{Advances in neural information processing systems}, 32, 2019.

\bibitem[Suh and Seo(2023)]{suh2023long}
Min-Kook Suh and Seung-Woo Seo.
\newblock Long-tailed recognition by mutual information maximization between latent features and ground-truth labels.
\newblock \emph{arXiv preprint arXiv:2305.01160}, 2023.

\bibitem[Tang et~al.(2020)Tang, Huang, and Zhang]{tang2020long}
Kaihua Tang, Jianqiang Huang, and Hanwang Zhang.
\newblock Long-tailed classification by keeping the good and removing the bad momentum causal effect.
\newblock \emph{Advances in Neural Information Processing Systems}, 33:\penalty0 1513--1524, 2020.

\bibitem[Van~Horn et~al.(2018)Van~Horn, Mac~Aodha, Song, Cui, Sun, Shepard, Adam, Perona, and Belongie]{van2018inaturalist}
Grant Van~Horn, Oisin Mac~Aodha, Yang Song, Yin Cui, Chen Sun, Alex Shepard, Hartwig Adam, Pietro Perona, and Serge Belongie.
\newblock The inaturalist species classification and detection dataset.
\newblock In \emph{Proceedings of the IEEE conference on computer vision and pattern recognition}, pages 8769--8778, 2018.

\bibitem[Wang et~al.(2021)Wang, Han, Wei, Zhang, and Wang]{wang2021contrastive}
Peng Wang, Kai Han, Xiu-Shen Wei, Lei Zhang, and Lei Wang.
\newblock Contrastive learning based hybrid networks for long-tailed image classification.
\newblock In \emph{Proceedings of the IEEE/CVF conference on computer vision and pattern recognition}, pages 943--952, 2021.

\bibitem[Wang et~al.(2017)Wang, Ramanan, and Hebert]{wang2017learning}
Yu-Xiong Wang, Deva Ramanan, and Martial Hebert.
\newblock Learning to model the tail.
\newblock \emph{Advances in neural information processing systems}, 30, 2017.

\bibitem[Xie et~al.(2017)Xie, Girshick, Doll{\'a}r, Tu, and He]{xie2017aggregated}
Saining Xie, Ross Girshick, Piotr Doll{\'a}r, Zhuowen Tu, and Kaiming He.
\newblock Aggregated residual transformations for deep neural networks.
\newblock In \emph{Proceedings of the IEEE conference on computer vision and pattern recognition}, pages 1492--1500, 2017.

\bibitem[Xuan and Zhang(2024)]{xuan2024decoupled}
Shiyu Xuan and Shiliang Zhang.
\newblock Decoupled contrastive learning for long-tailed recognition.
\newblock In \emph{Proceedings of the AAAI Conference on Artificial Intelligence}, pages 6396--6403, 2024.

\bibitem[Yang et~al.(2022)Yang, Chen, Li, Xie, Lin, and Tao]{yang2022inducing}
Yibo Yang, Shixiang Chen, Xiangtai Li, Liang Xie, Zhouchen Lin, and Dacheng Tao.
\newblock Inducing neural collapse in imbalanced learning: Do we really need a learnable classifier at the end of deep neural network?
\newblock \emph{Advances in Neural Information Processing Systems}, 35:\penalty0 37991--38002, 2022.

\bibitem[Zhang et~al.(2022)Zhang, Kuang, Chen, Liu, Wu, and Xiao]{zhang2022fairness}
Fengda Zhang, Kun Kuang, Long Chen, Yuxuan Liu, Chao Wu, and Jun Xiao.
\newblock Fairness-aware contrastive learning with partially annotated sensitive attributes.
\newblock In \emph{The Eleventh International Conference on Learning Representations}, 2022.

\bibitem[Zhang et~al.(2021)Zhang, Li, Yan, He, and Sun]{zhang2021distribution}
Songyang Zhang, Zeming Li, Shipeng Yan, Xuming He, and Jian Sun.
\newblock Distribution alignment: A unified framework for long-tail visual recognition.
\newblock In \emph{Proceedings of the IEEE/CVF conference on computer vision and pattern recognition}, pages 2361--2370, 2021.

\bibitem[Zhang et~al.(2023)Zhang, Kang, Hooi, Yan, and Feng]{zhang2023deep}
Yifan Zhang, Bingyi Kang, Bryan Hooi, Shuicheng Yan, and Jiashi Feng.
\newblock Deep long-tailed learning: A survey.
\newblock \emph{IEEE Transactions on Pattern Analysis and Machine Intelligence}, 45\penalty0 (9):\penalty0 10795--10816, 2023.

\bibitem[Zhong et~al.(2021)Zhong, Cui, Liu, and Jia]{zhong2021improving}
Zhisheng Zhong, Jiequan Cui, Shu Liu, and Jiaya Jia.
\newblock Improving calibration for long-tailed recognition.
\newblock In \emph{Proceedings of the IEEE/CVF conference on computer vision and pattern recognition}, pages 16489--16498, 2021.

\bibitem[Zhou et~al.(2017)Zhou, Lapedriza, Khosla, Oliva, and Torralba]{zhou2017places}
Bolei Zhou, Agata Lapedriza, Aditya Khosla, Aude Oliva, and Antonio Torralba.
\newblock Places: A 10 million image database for scene recognition.
\newblock \emph{IEEE transactions on pattern analysis and machine intelligence}, 40\penalty0 (6):\penalty0 1452--1464, 2017.

\bibitem[Zhou et~al.(2020)Zhou, Cui, Wei, and Chen]{zhou2020bbn}
Boyan Zhou, Quan Cui, Xiu-Shen Wei, and Zhao-Min Chen.
\newblock Bbn: Bilateral-branch network with cumulative learning for long-tailed visual recognition.
\newblock In \emph{Proceedings of the IEEE/CVF conference on computer vision and pattern recognition}, pages 9719--9728, 2020.

\bibitem[Zhu et~al.(2024)Zhu, Tang, Sun, and Zhang]{zhu2024generalized}
Beier Zhu, Kaihua Tang, Qianru Sun, and Hanwang Zhang.
\newblock Generalized logit adjustment: Calibrating fine-tuned models by removing label bias in foundation models.
\newblock \emph{Advances in Neural Information Processing Systems}, 36, 2024.

\bibitem[Zhu et~al.(2022)Zhu, Wang, Chen, Chen, and Jiang]{zhu2022balanced}
Jianggang Zhu, Zheng Wang, Jingjing Chen, Yi-Ping~Phoebe Chen, and Yu-Gang Jiang.
\newblock Balanced contrastive learning for long-tailed visual recognition.
\newblock In \emph{Proceedings of the IEEE/CVF Conference on Computer Vision and Pattern Recognition}, pages 6908--6917, 2022.

\end{thebibliography}
}

\end{document}